%% file: aaai-main.tex
\def\year{2020}\relax
\documentclass[letterpaper]{article} 
\usepackage{aaai20}  
\usepackage{times}  
\usepackage{helvet} 
\usepackage{courier}  
\usepackage[hyphens]{url}  
\usepackage{graphicx} 
\usepackage{graphicx}  
\usepackage{amsmath}
\usepackage{amssymb}
\title{Gated Fully Fusion for Semantic Segmentation}
\author{Xiangtai Li\textsuperscript{\rm 1}, Houlong Zhao\textsuperscript{\rm 2}, Lei Han\textsuperscript{\rm 3}, Yunhai Tong\textsuperscript{\rm 1}, Shaohua Tan\textsuperscript{\rm 1}, Kuiyuan Yang\textsuperscript{\rm 2}\\
\textsuperscript{\rm 1}School of EECS, Peking University
\textsuperscript{\rm 2}DeepMotion 
\textsuperscript{\rm 3}Tecent AI lab
\\
\{lxtpku, yhtong\}@pku.edu.cn, tan@cis.pku.edu.cn \\
\{houlongzhao, kuiyuanyang\}@deepmotion.ai \\
leihan.cs@gmail.com
}
 
\begin{document}

\maketitle

\input{0abstract.tex}

\input{1introduction.tex}
\input{2relatedwork.tex}
\input{3method.tex}
\input{4experiment.tex}
\input{5conclusion.tex}

{\small
\bibliographystyle{aaai}
\bibliography{egbib}
}

\section{More visualization of Gates}
\begin{figure*}
\centering
\includegraphics[width=0.8\linewidth]{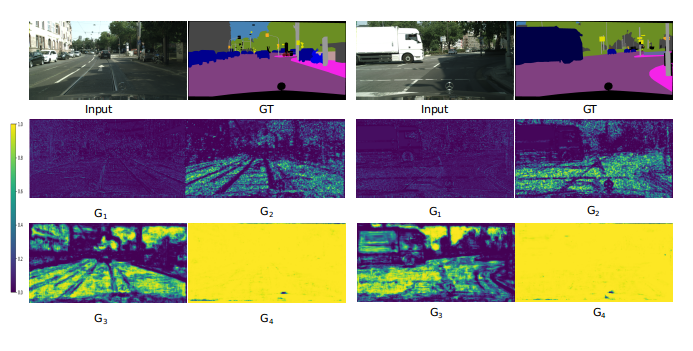}
\caption{
More visualization of learned gate maps on Cityscapes dataset. $G_i$ represents the output of $i_{th}$ layer's gate. It shows the gate control the information propagation. Best view in color and zoom in for detailed information.}
\label{fig:city_mask_vis}
\end{figure*}

\begin{figure*}
\centering
\includegraphics[width=0.9\linewidth]{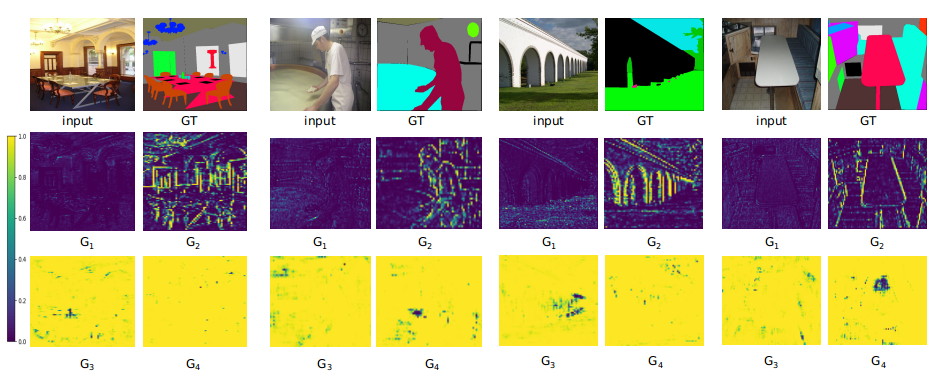}
\caption{More more visualization of learned gate maps on ADE20K dataset. $G_i$ represents the gate map of the $i$th layer. Best view in color and zoom in for detailed information.}
\label{fig:ade_mask_vis}
\end{figure*}

Here we give more visualization examples on Cityscapes~\cite{Cityscapes} and ADE20k~\cite{ADE20K} dataset shown in Fig~\ref{fig:city_mask_vis} and Fig~\ref{fig:ade_mask_vis} respectively.

\end{document}

%% file: 0abstract.tex
\begin{abstract}
Semantic segmentation generates comprehensive understanding of scenes through densely predicting the category for each pixel. High-level features from Deep Convolutional Neural Networks already demonstrate their effectiveness in semantic segmentation tasks, however the coarse resolution of high-level features often leads to inferior results for small/thin objects where detailed information is important. It is natural to consider importing low level features to compensate for the lost detailed information in high-level features.
Unfortunately, simply combining multi-level features suffers from the semantic gap among them. In this paper, we propose a new architecture, named Gated Fully Fusion (GFF), to selectively fuse features from multiple levels using gates in a fully connected way. Specifically, features at each level are enhanced by higher-level features with stronger semantics and lower-level features with more details, and gates are used to control the propagation of useful information which significantly reduces the noises during fusion. We achieve the state of the art results on four challenging scene parsing datasets including Cityscapes, Pascal Context, COCO-stuff and ADE20K.
\end{abstract}

%% file: 1introduction.tex
\section{Introduction}
\label{section:intro}
\begin{figure}
\centering
\includegraphics[width=1.0\linewidth]{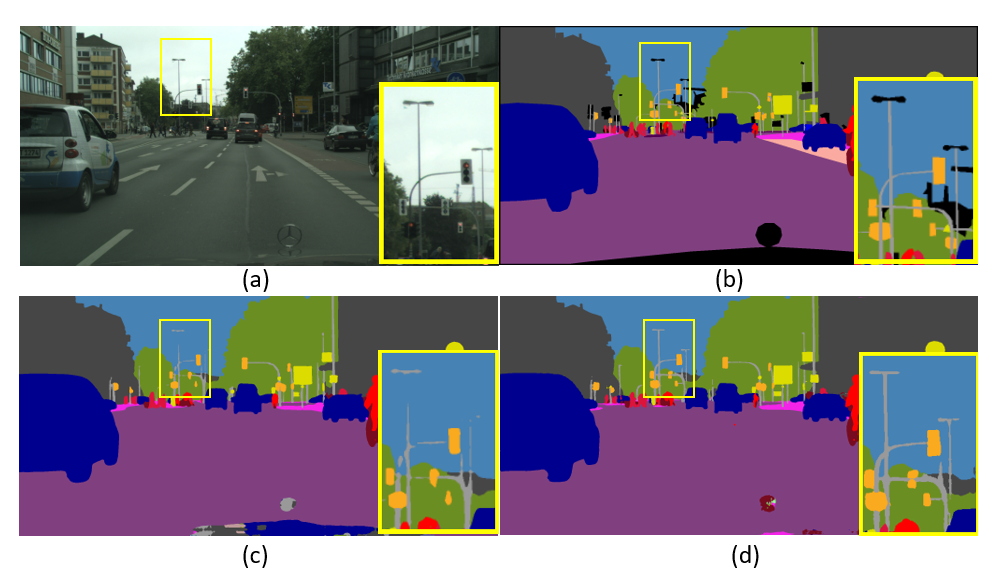}
\caption{Illustration of challenges in semantic segmentation.
(a) Input Image. 
(b) Ground Truth. 
(c) PSPNet result. 
(d) Our result. Our method performs much better on small patterns such as distant poles and traffic lights.}
\label{fig:intro}
\end{figure}

Semantic segmentation densely predicts the semantic category for every pixel in an image, such comprehensive image understanding is valuable for many vision-based applications such as medical image analysis~\cite{unet}, remote sensing~\cite{kampffmeyer2016semantic} and autonomous driving~\cite{xu2017end}. However, precisely predicting label for every pixel is challenging as illustrated in Fig.~\ref{fig:intro}, since pixels can be from tiny or large objects, far or near objects, and inside object or object boundary.

As a semantic prediction problem, the basic task of semantic segmentation is to generate \emph{high-level} representation for \emph{each} pixel, i.e., a \emph{high-level} and \emph{high-resolution} feature map. Given the ability of ConvNets in learning high-level representation from data, semantic segmentation has made much progress by leveraging such high-level representation. However, high-level representation from ConvNets is generated along lowering the resolution, thus high-resolution and high-level feature maps are distributed in two ends in a ConvNet.

To get a feature map that is both high-resolution and high-level, which is not readily available in a ConvNet, it is natural to consider fusing high-level feature maps from top layers and high-resolution feature maps from bottom layers. These feature maps are with different properties, that high-level feature map can correctly predict most of the pixels on large patterns in a coarse manner, which is widely used in the current semantic segmentation approaches, while low-level feature maps can only predict few pixels on small patterns. 

Thus, simply combining high-level feature maps and high-resolution feature maps will drown useful information in massive useless information, and cannot reach an informative high-level and high-resolution feature map. Therefore, an advanced fusion mechanism is required to collect information selectively from different feature maps. To achieve this, we propose Gated Fully Fusion (GFF) which uses gating mechanism, a kind of operation commonly used for information extraction from time series, to pixelwisely measure the usefulness of each feature vector, and control information propagation through gates accordingly. The principle of the gate at each layer is designed to either send out useful information to other layers or receive information from other layers when the information in the current layer is useless. Using gate to control information propagation, redundancies can also be effectively minimized in the network, allowing us to fuse multi-level feature maps in a fully-connected manner.  Fig~\ref{fig:intro} compares the results of GFF and PSPNet~\cite{pspnet}, where GFF can handle fine-level details such as poles and traffic lights in a much better way.

In addition, contextual information in large receptive field is also very important for semantic segmentation as proved by PSPNet~\cite{pspnet}, ASPP~\cite{deeplabv2} and DenseASPP~\cite{denseaspp}. Therefore, we also model contextual information after GFF to further improve the performance. Specifically, we propose a dense feature pyramid (DFP) module to encode context information into each feature map. DFP reuses the contextual information for each feature level and aims to enhance the context modeling part while GFF operates on the backbone network to capture more detailed information. Combining both components in a single end-to-end network, we achieve state-of-the-art results on four scene parsing datasets.

The main contributions of our work can be summarized as three points: Firstly, we propose Gated Fully Fusion to generate high-resolution and high-level feature map from multi-level feature maps, and Dense Feature Pyramid to enhance the semantic representation of multi-level feature maps. Secondly, detailed analysis with visualization of gates learned in different layers intuitively shows the information regulation mechanism in GFF. Finally, The proposed method is extensively verified on four standard semantic segmentation benchmarks including Cityscapes, Pascal Context, COCO-stuff and ADE20K, where our method achieves state-of-the-art performance on all four tasks. In particular, our models achieve \textbf{82.3\%} mIoU on Cityscapes test set with ResNet101 as backbone, \textbf{83.3\%} mIoU with WiderResNet as backbone which are trained \textbf{only on the fine labeled} data.


%% file: 2relatedwork.tex
\section{Related Work}
\label{section: related work}

\textbf{Context modeling}
Though high-level feature maps in ConvNets have shown promising results on semantic segmentation~\cite{fcn}, their receptive field sizes are still not large enough to capture contextual information for large objects and regions. Thus, context modeling becomes a practical direction in semantic segmentation. PSPNet~\cite{pspnet} uses spatial pyramid pooling to aggregate multi-scale contextual information. Deeplab series~\cite{deeplabv1,deeplabv2,deeplabv3} develop atrous spatial pyramid pooling (ASPP) to capture multi-scale contextual information by dilated convolutional layers with different dilation rates. Instead of parallel aggregation as adopted in PSPNet and Deeplab, Yang et al.~\cite{denseaspp} and Bilinski et al.~\cite{densedecoder} follow the idea of the dense connection~\cite{densenet} to encode contextual information in a dense way. In~\cite{largekernel}, factorized large filters are directly used to increase the receptive field size for context modeling. SVCNet~\cite{SVCNet} generates a scale and shape-variant semantic mask for each pixel to confine its contextual region.In PSANet~\cite{psanet}, contextual information is collected from all positions according to the similarities defined in a projected feature space. Similarly, DANet~\cite{DAnet}, CCNet~\cite{ccnet}, EMAnet~\cite{EMAnet} and ANN~\cite{annnet} use non-local style operator~\cite{Nonlocal} to aggregate information from the whole image based on pixel-to-pixel affinities.

\noindent
\textbf{Multi-level feature fusion}
 In addition to lack of contextual information, the top layer also lacks of fine detailed information. To address this issue, in FCN~\cite{fcn}, predictions from middle layers are used to improve segmentation for detailed structures, while hypercolumns~\cite{hypercol} directly combines features from multiple layers for prediction. The U-Net~\cite{unet} adds skip connections between the encoder and decoder to reuse low level features, ~\cite{Exfuse} improves U-Net by fusing high-level features into low-level features. Feature Pyramid Network (FPN)~\cite{fpn} uses the structure of U-Net with predictions from each level of the feature pyramid. DeepLabV3+~\cite{deeplabv3p} refines the decoder of its previous version by combing low-level features. In~\cite{msci} and ~\cite{ding2018context}, they proposed to locally fuse every two adjacent feature maps in the feature pyramid into one feature map until only one feature map is left. These fusion methods operate locally in the feature pyramid without awareness of the usefulness of all feature maps to be fused, which limits the propagation of useful features.

\noindent
\textbf{Gating mechanism}
  In deep neural networks, especially for recurrent networks, gates are commonly utilized to control information propagation. For example, LSTM~\cite{lstm} and GRU~\cite{gru} are two typical cases where different gates are used to handle long-term memory and dependencies. The highway network~\cite{highway} uses gates to make training deep network possible. To improve multi-task learning for scene parsing and depth estimation, PAD-Net~\cite{PAD-net} is proposed to use gates to fuse multi-modal features trained from multiple auxiliary tasks. DepthSeg\cite{depthseg} proposes depth-aware gating module which uses depth estimates to adaptively modify the pooling field size in high-level feature map. GSCNN~\cite{gated-scnn} uses gates to learn the precise boundary information by including another shape stream to encode edge feature into final representation. 
  
  Our method is related and inspired by the above methods, and differs from them in that multi-level feature maps are fused simultaneously through gating mechanism, and the resulting method surpasses the state-of-the-art approaches.

%% file: 3method.tex
\section{Method}

In this section, we first overview the basic setting of multi-level feature fusion and three baseline fusion strategies. Then, we introduce the proposed multi-level fusion module (GFF) and the whole network with the context modeling module (DFP).

\begin{figure}
\centering
\includegraphics[width=0.7\linewidth,height=50mm]{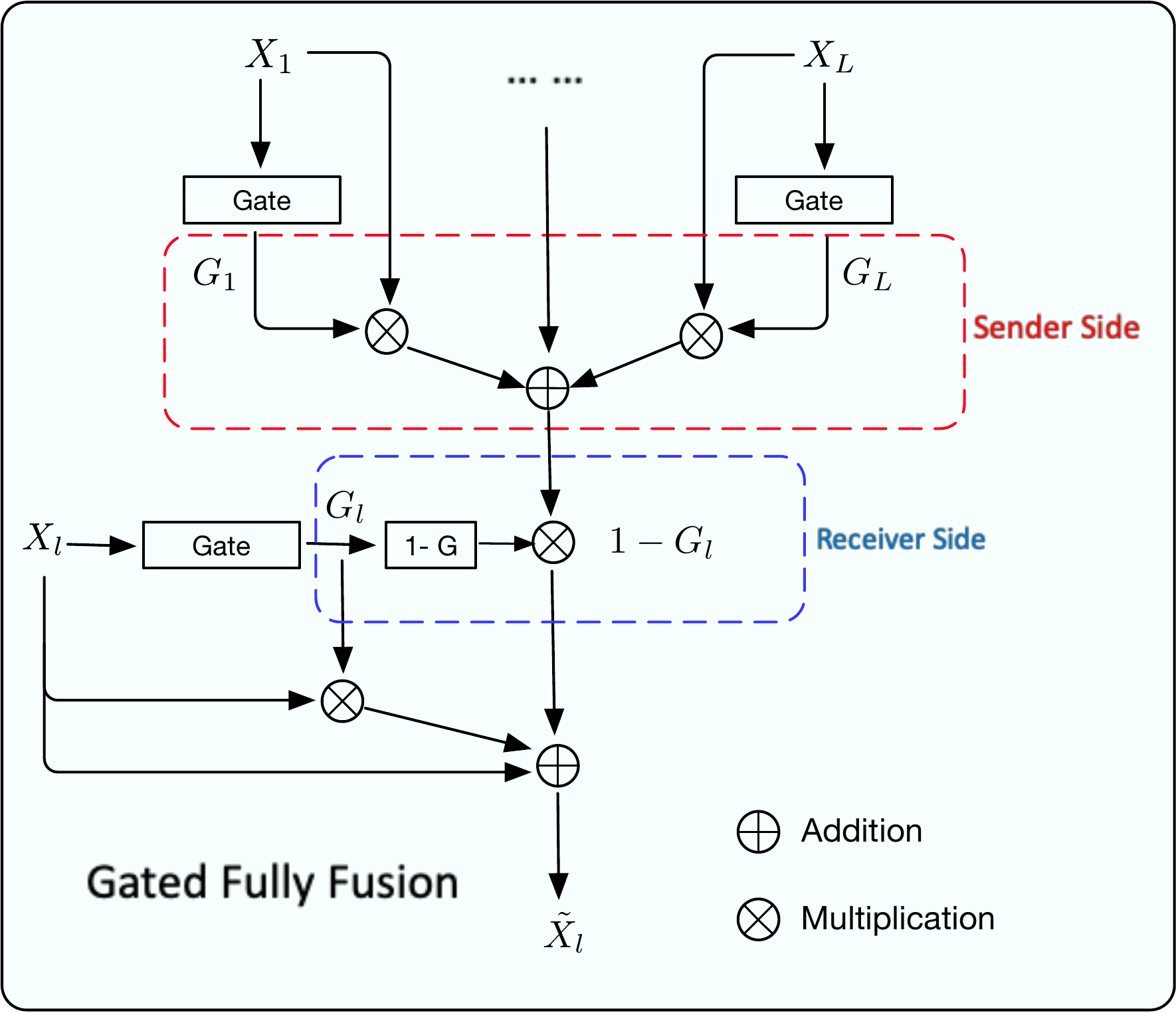}
\caption{\footnotesize The proposed gated fully fusion module,  where $G_l$ is the gate map generated from $X_l$, and features corresponding high gate values are allowed to send out and regions with low gate values are allowed to receive.}
\label{fig:fusion}
\end{figure}

\subsection{Multi-level Feature Fusion}
\label{sec:fusionreview}
  Given $L$ feature maps $\{X_{i}\in\mathbb{R}^{H_i\times W_i \times C_i}\}_{i=1}^L$ extracted from some backbone networks such as ResNet~\cite{resnet}, where feature maps are ordered by their depth in the network with increasing semantics but decreasing details, $H_i$, $W_i$ and $C_i$ are the height, width and number of channels of the $i$th feature map respectively, feature maps of higher levels are with lower resolution due to the downsampling operations, i.e., $H_{i+1} \leq H_i, W_{i+1} \leq W_i$. In semantic segmentation, the top feature map $X_L$ with $1/8$ resolution of the raw input image is mostly used for its rich semantics. The major limitation of $X_L$ is its low spatial resolution without detailed information, because the outputs need to be with the same resolution as the input image. In contrast, feature maps of low level from shallow layers are with high resolution, but with limited semantics. Intuitively, combining the complementary strengths of multiple level feature maps would achieve the goal of both high resolution and rich semantics, and this process can be abstracted as a fusion process $f$, i.e.,
  \begin{equation}
\begin{split}
    \{X_{1},X_{2} \cdots X_{L}\}  \overset{f}{\rightarrow} \{ \tilde{X_{1}}, \tilde{X_{2}} \cdots \tilde{X_{L}}\}
\end{split}
    \label{equ:fusion_process}
\end{equation}
where $\tilde{X}_l$ is the fused feature map for the $l$th level. To simplify the notations in following equations, bilinear sampling and $1\times 1$ convolution are ignored which are used to reshape the feature maps at the right hand side to let the fused feature maps have the same size as those at the left hand side. Concatenation is a straightforward operation to aggregate all the information in multiple feature maps, but it mixes the useful information with large amount of non-informative features. Addition is another simple way to combine feature maps by adding features at each position, while it suffers from the similar problem as concatenation. FPN~\cite{fpn} conducts the fusion process through a top-down pathway with lateral connections. The three fusion strategies can be formulated as,
    \begin{align}
       \textbf{Concat: } & \tilde{X_{l}} = \text{concat}(X_{1},...,X_L),\label{concat}\\
       \textbf{Addition: }& \tilde{X_{l}} = \sum_{i=1}^{L} X_{i},\label{equ:addition}\\
       \textbf{FPN: }  &\tilde{X}_{l} = \tilde{X}_{l+1} + X_{l} \text{, where } \tilde{X_{L}}=X_{L}.
    \end{align}
The problem of these basic fusion strategies is that feature maps are fused together without measuring the usefulness of each feature vector, and massive useless features are mixed with useful feature during fusion.
    
\begin{figure}
\centering
\includegraphics[width=1.0\linewidth]{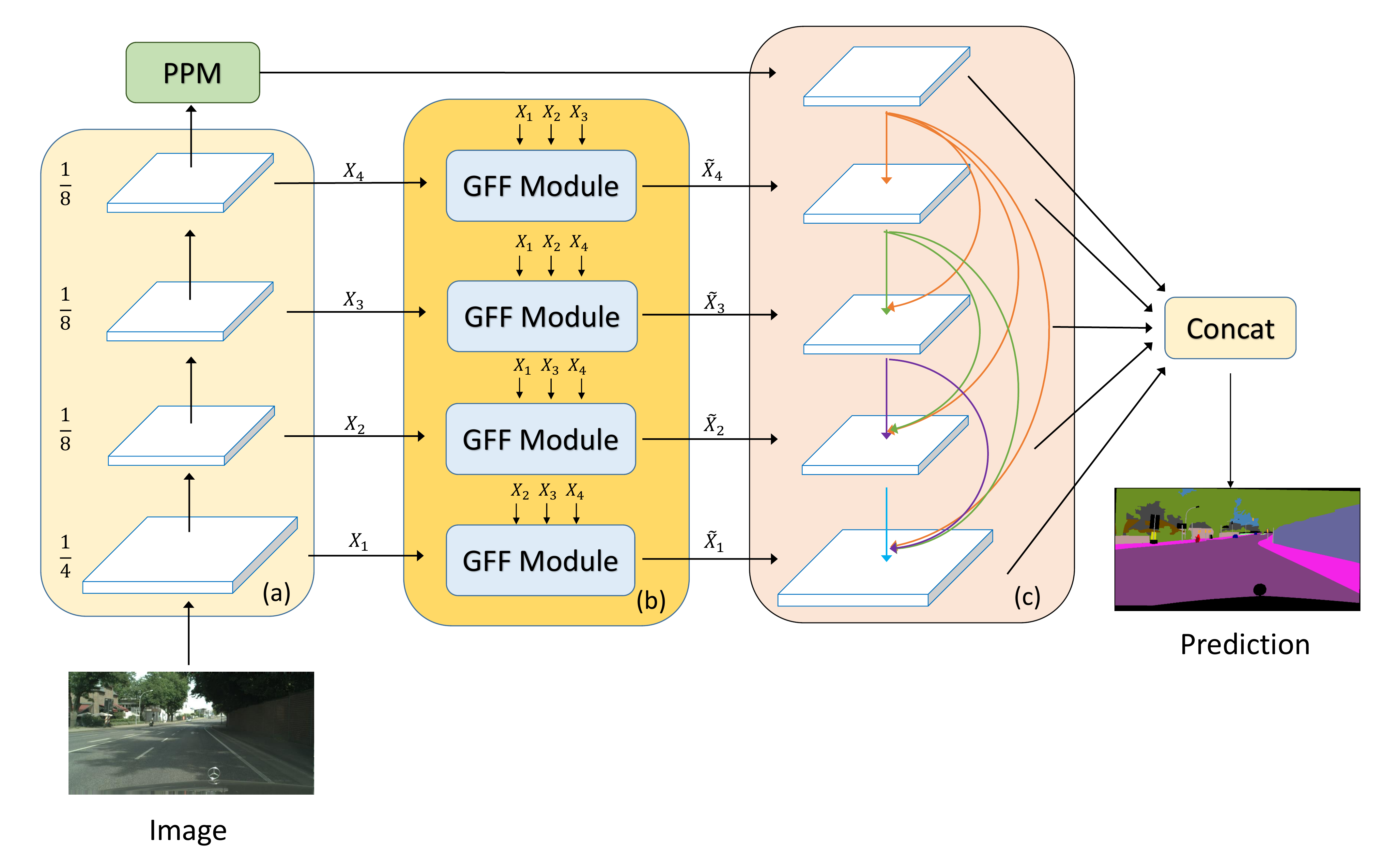}
\caption{\small{
Illustration of the overall architecture. 
(a) Backbone Network(e.g. ResNet~\cite{resnet}) with pyramid pooling module (PPM) \cite{pspnet} on the top.
The backbone provides a pyramid of features at different levels. (b), Feature pyramid through gated fully fusion (GFF) modules. The detail of the GFF module is illustrated in Fig~\ref{fig:fusion} . (c), Then the final features containing context information are obtained from a dense feature pyramid (DFP) module. Best view in color and zoom in.} }
\label{fig:overview}
\end{figure}

\subsection{Gated Fully Fusion}
\label{subsec:gated}
\textbf{GFF module design}: The basic task in multi-level feature fusion is to aggregate useful information together under interference of massive useless information. Gating is a mature mechanism to measure the usefulness of each feature vector in a feature map and aggregates information accordingly. In this paper, Gated Fully Fusion (GFF) is designed based on the simple addition-based fusion by controlling information flow with gates. Specifically, each level $l$ is associated with a gate map $G_l \in [0,1]^{H_l \times W_l}$. With these gate maps, the addition-based fusion is formally defined as
\begin{equation}
    \Tilde{X}_l = (1 + G_l) \cdot X_l + (1-G_l) \cdot \sum_{i=1, i \neq l}^{L}{G_i \cdot X_i,}\label{GFF}
\end{equation}
where $\cdot$ denotes element-wise multiplication broadcasting in the channel dimension, each gate map $G_l=\text{sigmoid}(\text{w}_i \ast X_i)$ is estimated by a convolutional layer parameterized with $\text{w}_i \in \mathbb{R}^{1\times 1 \times C_i}$. There are totally $\mathbf{L}$ gate maps where $\mathbf{L}$ equals to the number of feature maps. The detailed operation can be seen in Fig~\ref{fig:fusion}.

\noindent
\textbf{GFF involves duplex gating mechanism: } A feature vector at position $(x,y)$ from level $i$, (where $i\neq l$) can be fused to $l$ only when the value of $G_i(x,y)$ is large \textbf{and} the value of $G_l(x,y)$ is small, i.e., information is sent when level $i$ has the useful information that level $l$ is missing. Besides that useful information can be regulated to the right place through gates, useless information can also be effectively suppressed on both the sender and receiver sides, and information redundancy can be avoided because the information is only received when the current position has useless features. More visualization examples can be seen in experiments parts. 

\noindent
\textbf{Comparison with Other Gate module}: 
The work~\cite{ding2018context} also used gates for information control between adjacent layers. GFF differs in using gates to fully fuse features
from every level instead of adjacent levels, and richer information in all levels with large usability variance motivates us to design the duplex gating mechanism, which filters out useless information more effectively with gates at
both sides of the sender and receiver. Experimental results in the experiment section demonstrate the advantage of the proposed method.

\subsection{Dense Feature Pyramid}
Context modeling aims to encode more global information, and it is orthogonal to the proposed GFF becasue GFF is designed for backbone level. Therefore, we further design a module to encode more contextual information from outputs of both PSPNet~\cite{pspnet} and GFF. Motivated by that dense connections can strengthen feature propagation ~\cite{densenet}, we also densely connect the feature maps in a top-down manner starting from feature map outputted from the PSPNet, and high-level feature maps are reused multiple times to add more contextual information to low levels, which was found important in our experiments for correctly segmenting large pattern in objects. This process is shown as follows:
    \begin{equation}
        y_{i} = H_{i}([y_{0}, \tilde{X}_{1},...,\tilde{X}_{i-1}])
    \end{equation}
Consequently, the $j$-th feature pyramid receives the feature-maps of all preceding pyramids, $y_{0}$,$\tilde{X}_{1}$,...$\tilde{X}_{i-1}$ as input and outputs current pyramid $y_{i}$: where $x_{0}$ is the output of PSPNet and $\tilde{X}_{i}$ is the output of $i$-th GFF module. Fusion function $H_{i}$ is implemented by a single convolution layer. Since the feature pyramid is densely connected, we denote this module as Dense Feature Pyramid (DFP). 
The collections of DFP's outputs { $y_{i}$ } are used for final prediction. Both GFF and DFP can be plugged into existing FCNs for end-to-end training with only slightly extra computation cost.

\subsection{Network Architecture and Implementation}
Our network architecture is designed based on previous state-of-the-art network PSPNet~\cite{pspnet} with ResNet~\cite{resnet} as backbone for basic feature extraction, the last two stages in ResNet are modified with dilated convolution to make both strides to 1 and keep spatial information. Fig~\ref{fig:overview} shows the overall framework including both GFF and DFP. PSPNet forms the bottom-up pathway with backbone network and pyramid pooling module (PPM), where PPM is at the top to encode contextual information. Feature maps from last residual blocks in each stage of backbone are used as the input for GFF module, and all feature maps are reduced to 256 channels with $1\times 1$ convolutional layers. The output feature maps from GFF are further fused with two $3\times 3$ convolutional layers in each level before feeding into the DFP module. All convolutional layers are followed by batch normalization~\cite{batchnorm} and ReLU activation function. After DFP, all feature maps are concatenated for final semantic segmentation. Compared with the basic PSPNet, the proposed method only slightly increases the number of parameters and computations. The entire network is trained in an end-to-end manner driving by cross-entropy loss defined on the segmentation benchmarks. To facilitate the training process, an auxiliary loss together with the main loss are used to help optimization following~\cite{dsn}, where the main loss is defined on the final output of the network and the auxiliary loss is defined on the output feature map at stage3 of ResNet with weight of 0.4~\cite{pspnet}.

%% file: 4experiment.tex
\section{Experiment}
In this section, we analyze the proposed method on Cityscapes ~\cite{Cityscapes}dataset and report results on other datasets.

\begin{figure}
\centering
\includegraphics[width=1.0\linewidth]{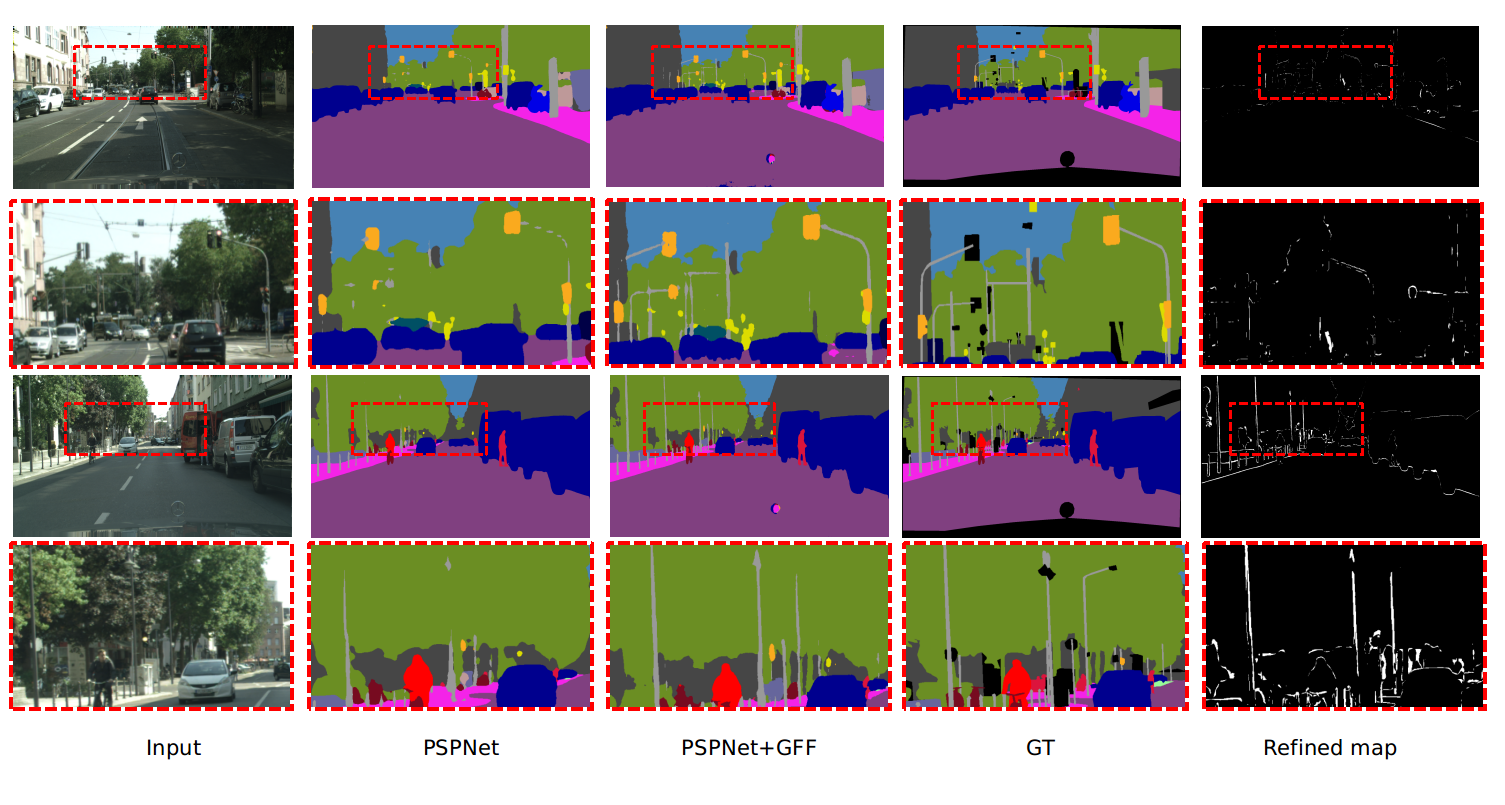}
\caption{
\small{
Visualization of segmentation results of two images using GFF and PSPNet. The first column shows two input images zoomed in regions marked with red dash rectangles. The second column shows results of PSPNet, and the third column shows results of using GFF. The fourth column lists the ground truth. The last column shows the refined parts by GFF. It shows that GFF can handle distant missing objects like poles, traffic lights and object boundaries. Best view in color.}}
\label{fig:gff_prediction_result}
\end{figure}

\subsection{Implementation Details}
Our implementation is based on PyTorch~\cite{pytorch}. The weight decay is set to 1e-4. Standard SGD is used for optimization, and ``poly'' learning rate scheduling policy is used to adjust learning rate, where initial learning rate is set to 1e-3 and decayed by 
$(1 - \frac{\text{iter}}{\text{total}\_\text{iter}})^{power}$ with $power = 0.9$. 
Synchronized batch normalization~\cite{context_encoding} is used. For Cityscapes, crop size of $864 \times 864$ is used, 100K training iterations with mini-batch size of 8 is carried for training. For ADE20K, COCO-stuff and Pascal Context, crop size of $512 \times 512$ is used (images with side smaller than the crop size are padded with zeros), 150K training iterations are used with mini-batch size of 16. As a common practice to avoid , data augmentation including random horizontal flipping, random cropping, random color jittering within the range of $[-10, 10]$, and random scaling in the range of $[0.75, 2]$ are used during training. 

\subsection{Experiments on Cityscapes Dataset}
\textbf{Cityscapes} is a large-scale dataset for semantic urban scene understanding. It contains 5000 fine pixel-level annotated images, which is divided into 2975, 500, and 1525 images for training, validation and testing respectively, where labels of training and validation are publicly released and labels of testing set are held for online evaluation. It also provides 20000 coarsely annotated images. 30 classes are annotated and 19 of them are used for pixel-level semantic labeling task. Images are in high resolution with the size of $1024 \times 2048$. The evaluation metric for this dataset is the mean Intersection over Union (mIoU).

\begin{figure}
\centering
\includegraphics[width=1.0\linewidth]{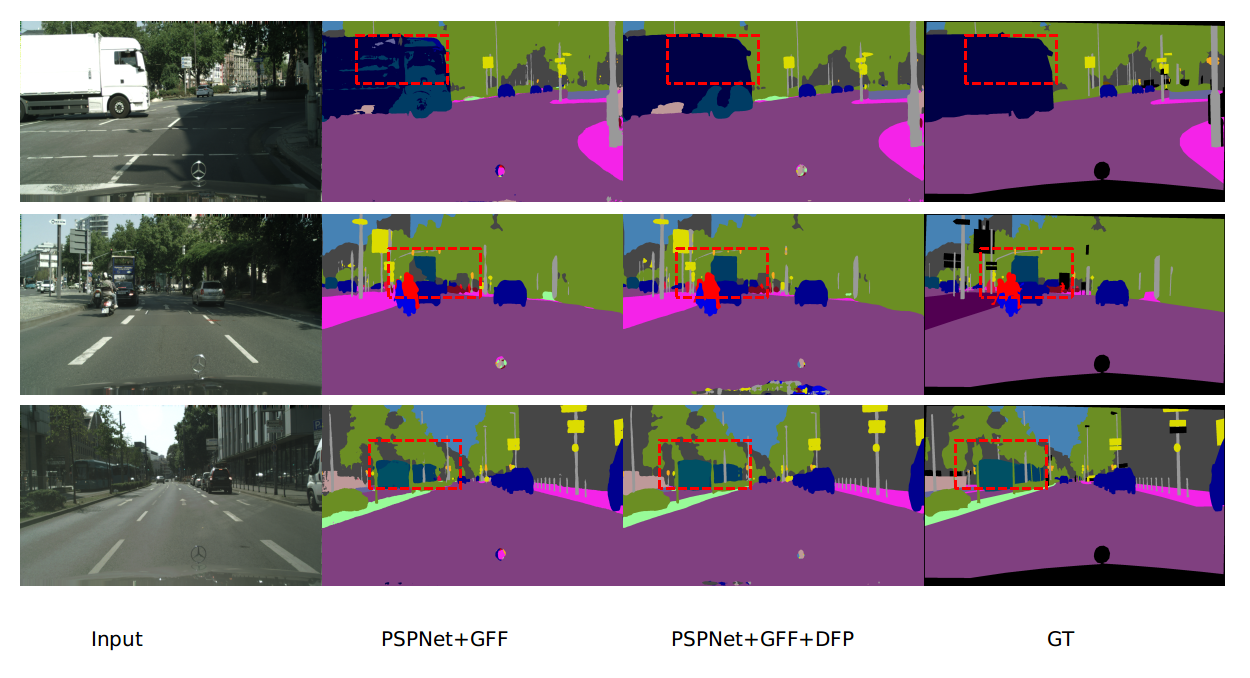}
\caption{DFP enhances segmentation results on large scale objects and generates more consistent results. Best view in color and zoom in.}
\label{fig:dfp_prediction_result}
\end{figure}

\noindent
\textbf{Strong Baseline} We choose PSPNet~\cite{pspnet} as our baseline model which achieved state-of-the-art performance for semantic segmentation. We re-implement PSPNet on Cityscapes and achieve similar performance with mIoU of 78.6\% on validation set. All results are reported by using sliding window crop prediction. 

\noindent
\textbf{Ablation Study on Feature Fusion Methods} First, we compare several methods introduced in Method part. To speed up the training process, we use weights from the trained PSPNet to initialize the parameters in each fusion method. We use train-fine data for training and report performance on validation set. For fair comparison with concatenation and addition, we also reduce the channel dimension of feature maps to 256 and use two $3 \times 3$ convolutional layers to refine the fused feature map. As for FPN, we implement the original FPN for semantic segmentation following~\cite{fpn_slide} and we add it to PSPNet. Note that FPN based on PSPNet fuses 5 feature maps, where one is context feature map from pyramid pooling module and others are from the backbone. 

All the results are shown in Table~\ref{tab:exam_cityscapes}. As expected, concatenation and addition only slightly improve the baseline, and FPN achieves the best performance among the three base fusion methods, while the proposed GFF obtains even  more improvement with mIoU of $80.4\%$. Since GFF is a gated version of addition-based fusion, the results demonstrate the effectiveness of the used gating mechanism. For further comparison, we also add the proposed gating mechanism into top-down pathway of FPN and observe slight improvement, which is reasonable since most high-level features are useful for low levels. This demonstrates the advantage of fully fusing multi-level feature maps, and the importance of gating mechanism especially during fusing low-level features to high levels. Fig~\ref{fig:gff_prediction_result} shows results after using GFF, where the accuracies of predictions for both distant objects and object boundaries are significantly improved.

\begin{table}[!t]
    \centering
    \begin{tabular}{l|l}
    \hline
    Method & mIoU(\%) \\
    \hline
    PSPNet(Baseline) & 78.6 \\
    PSPNet + Concat & 78.8 (0.2$\uparrow$)\\
    PSPNet + Addition & 78.7 (0.1$\uparrow$)\\
    PSPNet + FPN & 79.3 (0.7$\uparrow$) \\
    \hline
    PSPNet + Gated FPN & 79.4 (0.8 $\uparrow$) \\
    \hline \hline
    PSPNet + GFF & 80.4 (1.8$\uparrow$) \\
    \hline
    \end{tabular}
     \caption{Comparison experiments on Cityscapes validation set, where PSPNet serves as the baseline method.}
     \label{tab:exam_cityscapes}
\end{table}

\noindent
\textbf{Ablation Study for Improvement Strategies} We perform two strategies to further boost the performance of our model:, (1) DFP: Dense Feature Pyramid is used after the output of GFF module; and (2) MS: multi-scale inference is adopted, where the final segmentation map is averaged from segmentation probability maps with scales $\{0.75, 1, 1.25, 1.5, 1.75\}$ for evaluation. Experimental results are shown in Table~\ref{tab:improve}, and DFP further improves the performance by $0.8\%$ mIoU. Fig.~\ref{fig:dfp_prediction_result} shows several visual comparisons, where DFP generates more consistent segmentation inside large objects and demonstrates the effectiveness in using contextual information for resolving local ambiguities. With multi-scale inference, our model achieves $81.8\%$ mIoU, which significantly outperforms previous state-of-the-art model DeepLabv3+ ($79.55\%$ on Cityscapes validation set) by \textbf{2.25\%}.

\noindent
\textbf{Ablation Study for other architectures} we also perform experiments two different backbone architectures~\cite{denseaspp}. One is another strong baseline and the other is PSPNet with lightweight backbone. Results are shown in table~\ref{tab:more_bases}. It shows that both GFF and DFP show their generality on improving model results. In particular, resnet18 based PSPnet improve \textbf{5.9\% point} from the baseline.

\noindent
\textbf{Computation Cost} In Table~\ref{tab:GFlops}, we also study the computational cost of using our modules, where our method spends 7.7\% more computational cost and 6.3\% more parameters compared with the baseline PSPNet which indicates our method can be easily plugged in existing state-of-art segmentation methods with little extra computation cost.

\begin{table}[!t]
    \centering
    \footnotesize
    \begin{tabular}{l|l}
    \hline
    Method & mIoU(\%) \\
    \hline
    PSPNet(Baseline) & 78.6 \\
    \hline
    PSPNet + GFF & 80.4 (1.8$\uparrow$) \\
    \hline
    \hline
    PSPNet + GFF + DFP & 81.2 (2.6$\uparrow$) \\
    \hline
    PSPNet + GFF + DFP + MS & 81.8 (3.2$\uparrow$) \\ 
    \hline
    \end{tabular}
     \caption{Comparison experiments on Cityscapes validation set, where PSPNet serves as the baseline method.}
     \label{tab:improve}
\end{table}

\begin{table}[!t]
    \centering
    \footnotesize
    \setlength{\tabcolsep}{3.4 pt}
    \begin{tabular}{l|l|l|l}
    \hline
    Method & mIoU(\%) & FLOPS(G) & Params(M) \\
    \hline
    PSPNet(Baseline) & 78.6 & 580.1 & 65.6\\
    \hline
    PSPNet + GFF & 80.4 & 600.1 & 69.7 \\
    \hline
    PSPNet + GFF + DFP & 81.2 &  625.5 & 70.5\\ 
    \hline
    \end{tabular}
     \caption{Computational cost comparison, where PSPNet serves as the baseline with image of size $512\times512$ as input.}
     \label{tab:GFlops}
\end{table}

	\begin{table}[!t]
		\centering
		\footnotesize
		\begin{tabular}{l|l|l}
			\hline
			Method &  Backbone & mIoU\% \\
			\hline
			DenseASPP & DenseNet121 & 78.9 \\
			DenseASPP+GFF & DenseNet121 & 80.1(1.2$\uparrow$) \\
			DenseASPP+GFF+DFP & DenseNet121 & 80.9(2.0$\uparrow$) \\
			\hline
            PSP & ResNet18 &  73.0 \\
            PSP + GFF & ResNet18 & 76.6 (3.6$\uparrow$)\\
            PSP + GFF+DFP & ResNet18 & 78.9 (5.9$\uparrow$) \\
			\hline
		\end{tabular}
		\caption{Ablation study on two different models, where mIoU is evaluated on Cityscapes validation set.}
		\label{tab:more_bases}
	\end{table}

\begin{table}[!t]
\centering
\footnotesize
\setlength{\tabcolsep}{1.0 pt}
\begin{tabular}{l|l|l}
\hline
Method & Backbone & mIoU(\%)  \\
\hline
PSPNet~\cite{pspnet}\textdagger  & ResNet101 &  78.4 \\ 
PSANet~\cite{psanet}\textdagger & ResNet101 &  78.6 \\
\hline
GFFNet(Ours)\textdagger & ResNet101 &  \textbf{80.9} \\
\hline
AAF~\cite{aaf}\textdaggerdbl  & ResNet101 &  79.1 \\ 
PSANet~\cite{psanet}\textdaggerdbl & ResNet101 &  80.1 \\ 
DFN~\cite{dfn}\textdaggerdbl & ResNet101 &  79.3 \\ 
DepthSeg~\cite{depthseg}\textdaggerdbl & ResNet101 &  78.2 \\ 
DenseASPP~\cite{denseaspp} \textdaggerdbl & DenseNet161 & 80.6 \\
SVCNet~\cite{SVCNet} \text\textdaggerdbl & ResNet101 & 81.0 \\
DANet~\cite{DAnet} \textdaggerdbl & ResNet101 & 81.5 \\ 
\hline
GFFNet(Ours)\textdaggerdbl & ResNet101 & \textbf{82.3} \\
\hline
\end{tabular}
\caption{State-of-the-art comparison experiments on Cityscapes test set. \textdagger  means only using the train-fine dataset. \textdaggerdbl means both the train-fine and val-fine data are used.}
\label{tab:cityscapes_results}
\end{table}

\begin{table}[!ht]
\centering
\footnotesize
\setlength{\tabcolsep}{0.6 pt}
\begin{tabular}{l|c|c|c}
\hline
Method & Backbone & mIoU($\%$) & Pixel Acc.(\%) \\
\hline
RefineNet~\cite{refinenet}& ResNet101 & 40.20 & -  \\ 
PSPNet~\cite{pspnet} & ResNet101 &  43.29 & 81.39\\ 
PSANet~\cite{psanet} & ResNet101 &  43.77 & 81.51\\ 
EncNet~\cite{context_encoding}  & ResNet101 & 44.65 & 81.69
\\ 
GCUNet~\cite{beyond_grids} & ResNet101 & 44.81 & 81.19\\
\hline
GFFNet(Ours) & ResNet101 & \textbf{45.33} & \textbf{82.01} \\
\hline
\end{tabular}
\caption{\small{State-of-the-art comparison experiments on ADE20K validation set. Our models achieve top performance measured by both mIoU and pixel accuracy.} }
\label{tab:ade20k_res}
\end{table}

\begin{table*}[t]
	\footnotesize
	\setlength{\tabcolsep}{0.8 pt}
		\begin{tabular}{ l | c | c | c | c | c | c | c | c | c | c | c| c| c| c| c| c| c| c| c | c}
		    \hline
			Method & road & swalk & build & wall & fence & pole & tlight & sign & veg. & terrain & sky & person & rider & car & truck & bus & train & mbike & bike & mIoU \\
			\hline
			PSPNet~\cite{pspnet} & 98.6 & 86.2 & 92.9 & 50.8 & 58.8 & 64.0 & 75.6 & 79.0 & 93.4 & 72.3 & 95.4 & 86.5 & 71.3 & 95.9 & 68.2 & 79.5 & 73.8 & 69.5 & 77.2 & 78.4\\
			AAF~\cite{aaf} & 98.5 & 85.6 & 93.0 & 53.8 & 58.9 & 65.9 & 75.0 & 78.4 & 93.7 &
			72.4 & 95.6 & 86.4 & 70.5 & 95.9 & 73.9 & 82.7 & 76.9 & 68.7 & 76.4 & 79.1 \\
			DenseASPP~\cite{denseaspp}  & 98.7 & 87.1 & 93.4 & \textbf{60.7} & 62.7 & 65.6 & 74.6 & 78.5 & 93.6 & 72.5 & 95.4 & 86.2 & 71.9 & 96.0 & 78.0 & 90.3 & 80.7 & 69.7 & 76.8 & 80.6 \\
			DANet~\cite{DAnet} & 98.6 & 87.1 & 93.5 & 56.1 & 63.3 & 69.7 & 77.3 & 81.3 & 93.9 & \textbf{72.9} & 95.7 & 87.3 & 72.9 & 96.2 & 76.8 & 89.4 & \textbf{86.5} & \textbf{72.2} & 78.2 & 81.5 \\
			\hline
		GFFNet(Ours) & \textbf{98.7} & \textbf{87.2} & \textbf{93.9} & 59.6 & \textbf{64.3} & \textbf{71.5} & \textbf{78.3} & \textbf{82.2} & \textbf{94.0} & 72.6 & \textbf{95.9} & \textbf{88.2} & \textbf{73.9} & \textbf{96.5} & \textbf{79.8} & \textbf{92.2} & 84.7 & 71.5 & \textbf{78.8} & \textbf{82.3} \\
		\hline
		\end{tabular}
	\caption{\small{
	Per-category results on Cityscapes test set. Note that all the models are trained with \textbf{only fine annotated data}. Our method outperforms existing approaches on 15 out of 19 categories, and achieves 82.3\% mIoU. }
	}
	\label{tab:cityscapes_results_detail_fine}
\end{table*}

\noindent
\textbf{Comparison to the State-of-the-Art} 
As a common practice toward best performance, we average the predictions of multi-scaled images for inference. For fair comparison, all methods are only trained using fine annotated dataset and evaluated on test set by the evaluation server. Table~\ref{tab:cityscapes_results} summarizes the comparisons, our method achieves 80.9\% mIoU by only using train-fine dataset and outperforms PSANet~\cite{psanet} by \textbf{2.3\%}. By fine-tuning the model on both train-fine and val-fine datasets, our method achieves the best mIoU of \textbf{82.3\%}. Detailed per-category comparisons are reported in Table~\ref{tab:cityscapes_results_detail_fine}, where our method achieves the highest IoU on 15 out of 19 categories, and large improvements are from small/thin categories such as pole, street light/sign, person and rider. 

Moreover, we further apply our methods on state-of-the-art method Deeplabv3+~\cite{deeplabv3p} with more stronger backbone Wider-ResNet~\cite{wide_resnet} pretrained on Mapillary~\cite{mapillary} dataset which shares the same setting with GSCNN~\cite{gated-scnn}. Table~\ref{tab:cityscapes_results_detail_large} shows the detailed results of our methods with previous state-of-art methods which also use corase data annotations. For both experiments, \textbf{we don't use coarse data}. More detailed analysis will be given by gate visualization.

\begin{table*}[t!]
\centering 
\addtolength{\tabcolsep}{-2pt}
\resizebox{\textwidth}{!}{
  \begin{tabular}{l|c| c|c|c|c|c|c|c|c|c|c|c|c|c|c|c|c|c|c|c|c}
  \hline
 Method & Coarse &  road & swalk & build. & wall & fence & pole & tlight & sign & veg & terrain & sky & person & rider & car & truck & bus & train & motor & bike & mIoU \\
\hline
        PSP-Net \cite{pspnet} & \checkmark &98.7 & 86.9 &  93.5 & 58.4 & 63.7 &  67.7 & 76.1 & 80.5 & 93.6 & 72.2  & 95.3  & 86.8 & 71.9 & 96.2 & 77.7 & 91.5 & 83.6 & 70.8 & 77.5 & 81.2 \\ 
         DeepLabV3 \cite{deeplabv3} & \checkmark& 98.6 & 86.2 & 93.5 & 55.2 & 63.2 & 70.0& 77.1 & 81.3 & 93.8    & 72.3&   95.9  &  87.6  & 73.4& 96.3 & 75.1 &  90.4  & 85.1 & 72.1 & 78.3 & 81.3 \\
         DeepLabV3+ \cite{deeplabv3p} & \checkmark &98.7 & 87.0  & 93.9 & 59.5 & 63.7 & 71.4 &78.2 & 82.2& 94.0& 73.0 & 95.8&88.0&  73.3 & 96.4  &  78.0 & 90.9 & 83.9 & 73.8 & 78.9  & 81.9 \\
         AutoDeepLab-L \cite{auto-deeplab} & \checkmark & \bf{98.8} & 87.6  & 93.8  & 61.4  & 64.4  & 71.2 & 77.6 & 80.9 & 94.1 & 72.7 & 96.0 & 87.8 &  72.8  & 96.5  & 78.2 & 90.9  & 88.4  & 69.0  &  77.6  &  82.1 \\
         DPC \cite{DPC} & \checkmark & 98.7 & 87.1 & 93.8 & 57.7 & 63.5 & 71.0 & 78.0 & 82.1 &94.0 & 73.3 & 95.4 & 88.2 &  \bf{74.5}  & \bf{96.5}  &  \bf{81.2} & \bf{93.3}  & \bf{89.0}  & \bf{74.1}  & 79.0   &   82.7 \\
   \hline       
    G-SCNN~\cite{gated-scnn} &  & 98.7 & 87.4 & 94.2 & 61.9 & \bf{64.6} & \bf{72.9} & \bf{79.6} & \bf{82.5} & \bf{94.3} & \bf{74.3}& \bf{96.2} & 88.3 & 74.2 & 96.0 & 77.2& 90.1 & 87.7& 72.6& \bf{79.4} & 82.8 \\
    GFFNet(ours) & & \bf{98.8} & \bf{87.9} & \bf{94.3} & \bf{64.7} & 65.8 & 71.9 & 78.9 & 82.4 & 94.2 & \bf{74.3} & 96.1 & \bf{88.4} & \bf{74.9} & \bf{96.5} & 79.2 & 92.8 & \bf{90.2} & 73.4 & 79.1 & \bf{83.3}\\
\hline
\end{tabular}
}
\vspace{-3mm}
\caption{ 
\small Per-category results on Cityscapes test set. Note that our methods and G-SCNN are trained with \textbf{only fine annotated data}.
} 
\label{tab:cityscapes_results_detail_large}
\end{table*}

\begin{figure}
\centering
\includegraphics[width=0.9\linewidth]{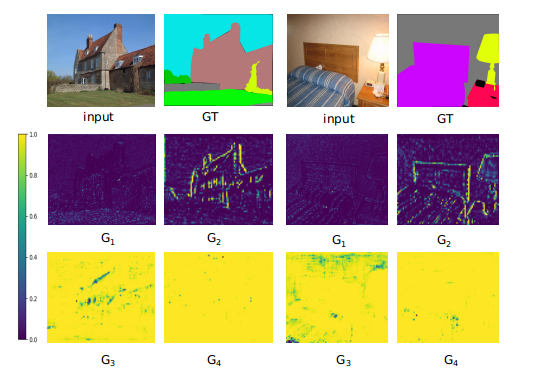}
\caption{Visualization of learned gate maps on ADE20K dataset. $G_i$ represents the gate map of the $i$th layer. Best view in color and zoom in for detailed information.}
\label{fig:ade_mask_vis}
\end{figure}

\subsection{Visualization of Gates}

\begin{figure}
\centering
\includegraphics[width=1.0\linewidth]{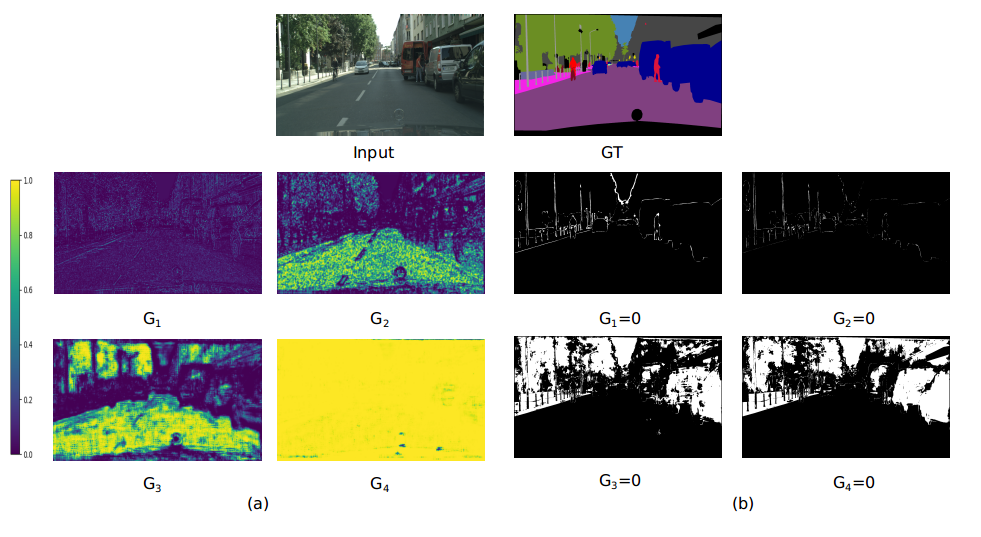}
\caption{
(a) Visualization of learned gate maps on Cityscapes dataset, where $G_i$ represents the gate map of the $i$th layer. (b) Wrongly classified pixels are highlighted after setting $G_i$ to 0 comparing with using original gate values.
Best view in color and zoom in for detailed information.}
\label{fig:city_mask_vis}
\end{figure}

    In this section, we visualize what gates have learned and analyze how gates control the information propagation. Fig~\ref{fig:ade_mask_vis} shows the gates learned from ADE20K and Fig~\ref{fig:city_mask_vis}(a) shows the gates learned from Cityscapes respectively. For each input image, we show the learned gate map of each level. As expected, we find that the higher-level features (e.g., $G_{3}$, $G_{4}$) are more useful for large structures with explicit semantics, while the lower-level features (e.g., $G_{1}$ and $G_{2}$) are mainly useful for local details and boundaries.
    
    Functionally, we find that the higher level features always spread information to other layers and only receive sparse feature signals. For example, the gate from stage 4 (in $G_{4}$ of Fig~\ref{fig:city_mask_vis}) shows that almost all pixels are of high-confidence. Higher-level features cover a large receptive field with fewer details, and they can provide a ground scope of the main semantics.
    
    In contrast, the lower level layers prefer to receive information while only spreading a few sparse signals. This verifies that lower level representations generally vary frequently along the spatial dimension and they require additional features as semantic supplement, while a benefit is that lower features can provide precise information for details and object boundaries ($G_{2}$ in Fig~\ref{fig:ade_mask_vis} and $G_{1}$ in Fig~\ref{fig:city_mask_vis}(a)).
    
    To further verify the effectiveness of the learned gates, we set the value of each gate $G_{i}$ to zero and compare the segmentation results with learned gate values.
    Fig~\ref{fig:city_mask_vis} (b) shows the comparison results, where wrongly predicted pixels after setting $G_{i}$ to zero are highlighted. Information through $G_{1}$ and $G_{2}$ is mainly help for object boundaries, while information through $G_{3}$ and $G_{4}$ is mainly help for large patterns such as cars. Additional visualization examples for the gates can be found in the supplementary materials.
    
\subsection{Results on Other Datasets}
\begin{figure}
\centering
\includegraphics[width=0.8\linewidth]{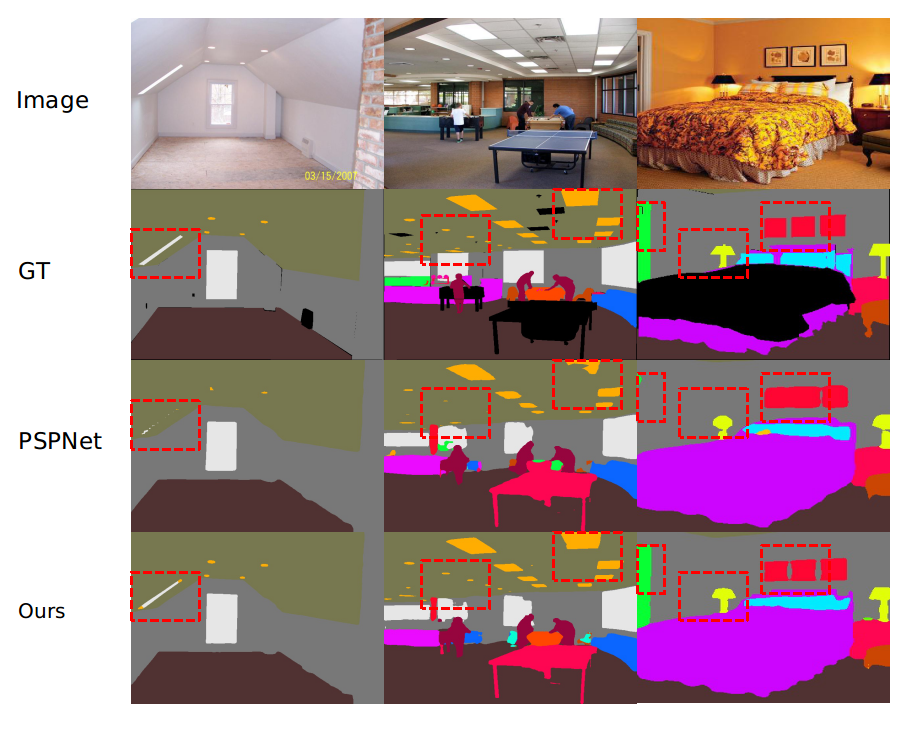}
\caption{Visualization results on ADE20K validation dataset (ResNet101 as backbone). Comparing with PSPNet, our method captures more detailed information, and finds missing small objects (e.g., lights in first two examples) and generates ``smoother'' on object boundaries (e.g., figures on the wall in last example). Best view in color.}
\label{fig:ade_vis_res}
\end{figure}

\textbf{ADE20K} is a challenging scene parsing dataset annotated with 150 classes, and it contains 20K/2K images for training and validation. Images in this dataset are from more different scenes with more small scale objects, and are with varied sizes including max side larger than 2000 and min side smaller than 100. Following the standard protocol, both mIoU and pixel accuracy evaluated on validation set are used as the performance metrics.In table\ref{tab:ade20k_res}, with backbone ResNet101, our method outperforms state-of-the-art methods with considerable margin in terms of both mIoU and pixel accuracy. Several visual comparison results are shown in Fig~\ref{fig:ade_vis_res}, where our method performs much better at details and object boundaries.

    \begin{table}[!t]
    	\centering
    	\footnotesize
    	\setlength{\tabcolsep}{2.0 pt}
    	\begin{tabular}{l|l|l}
    		\hline
    		Method & BackBone & mIoU(\%) \\
    		\hline
    		EncNet~\cite{context_encoding}  & ResNet-50   &49.0  \\
    		DANet~\cite{DAnet}&ResNet-50  & 50.1   \\
    		\hline
    		GFFNet(Ours) & ResNet50 & \textbf{51.0} \\
    		\hline
    	    PSPNet~\cite{pspnet}&ResNet-101   &47.8  \\
            EncNet~\cite{context_encoding}  & ResNet-101   &51.7  \\
            CCLNet~\cite{ding2018context} & ResNet101 & 51.6 \\
            DANet~\cite{DAnet}& ResNet-101  & 52.6   \\
            SVCNet~\cite{SVCNet} & ResNet-101 & 53.2 \\
    		\hline
    		GFFNet(Ours) & ResNet101 & \textbf{54.2} \\
    		\hline
    	\end{tabular}
    	\caption{Results on Pascal Context testing set. }
    	\label{tab:pascal_context}
    \end{table}

    \begin{table}[!t]
		\centering
		\footnotesize
		\setlength{\tabcolsep}{2.0 pt}
		\begin{tabular}{l|l|l}
			\hline
			Method & BackBone & mIoU(\%) \\
			\hline
			RefineNet~\cite{refinenet} & ResNet101 & 33.6\\
			DSSPN~\cite{DSSPN} & ResNet101 & 36.2 \\
			CCLNet~\cite{ding2018context} & ResNet101 & 35.7 \\
			\hline
			GFFNet(Ours) & ResNet101 & \textbf{39.2} \\
			\hline
		\end{tabular}
		\caption{Results on COCO stuff testing set. }
		\label{tab:coco_stuff}
	\end{table}

\noindent
\textbf{Pascal Context}~\cite{pcontext-data} provides pixel-wise segmentation annotation for 59 classes. There are 4998 training images and 5105 testing images. The results are shown in Table~\ref{tab:pascal_context}. Our method achieves the state-of-the-art results on both ResNet50 and ResNet101 backbone and outperforms the existing methods by
a large margin. 

\noindent
\textbf{COCO Stuff}~\cite{coco_stuff} contains 10000 images from Microsoft
COCO dataset ~\cite{COCO_dataset}, out of which 9000 images are for
training and 1000 images for testing. This dataset contains 171 categories including objects and stuff annotated to each pixel. The results of COCO Stuff are shown in Table~\ref{tab:coco_stuff}. Our method outperforms the existing methods and achieves top performance.

%% file: 5conclusion.tex
\section{Conclusion}
\label{conclusion}
In this work, we propose Gated Fully Fusion (GFF) to fully fuse multi-level feature maps controlled by learned gate maps. The novel module bridges the gap between high resolution with low semantics and low resolution with high semantics. We explore the proposed GFF for the task of semantic segmentation and achieve new state-of-the-art results four challenging scene parsing dataset. In particular, we find that the missing low-level features can be fused into each feature level in the pyramid, which indicates that our module can well handle small and thin objects in the scene. 